\documentclass[a4paper]{spie}  
\usepackage[]{graphicx,epsfig}

\title{Colonoscope tracking method based on \\shape estimation network}

\author{Masahiro ODA\supit{a}, Holger R. Roth\supit{a}, Takayuki KITASAKA\supit{b}, Kazuhiro FURUKAWA\supit{c}, \\Ryoji MIYAHARA\supit{d}, Yoshiki HIROOKA\supit{c}, Nassir NAVAB\supit{e}, and Kensaku MORI\supit{a,f}
\skiplinehalf
\supit{a}Graduate School of Informatics, Nagoya University, \\
Furo-cho, Chikusa-ku, Nagoya, Aichi, 464-8601, Japan; \\
\supit{b}School of Information Science, Aichi Institute of Technology, \\
1247 Yachigusa, Yagusa-cho, Toyota, Aichi, 470-0392, Japan; \\
\supit{c}Department of Endoscopy, Nagoya University Hospital, \\
65 Tsurumai-cho, Syouwa-ku, Nagoya, Aichi, Japan; \\
\supit{d}Dept. of Gastro. and Hepato., Nagoya University Graduate School of Medicine, \\
65 Tsurumai-cho, Syouwa-ku, Nagoya, Aichi, Japan; \\
\supit{e}Technical University of Munich, \\
Boltzmannstr. 3, 85748 Garching bei M\"{u}nchen, Germany; \\
\supit{f}Research Center for Medical Bigdata, National Institute of Informatics, \\
2-1-2 Hitotsubashi, Chiyoda-ku, Tokyo, 101-8430, Japan
}

\authorinfo{Further author information: 
(Send correspondence to M. Oda)
\\M. Oda: E-mail: moda@mori.m.is.nagoya-u.ac.jp, Telephone: +81 (0)52 789 5688
\\T. Kitasaka: E-mail: kitasaka@aitech.ac.jp, Telephone: +81 (0)565 48 8121\\ K. Mori: E-mail: kensaku@is.nagoya-u.ac.jp, Telephone: +81 (0)52 789 5689
}

\begin{document} 
  \maketitle 

\begin{abstract}
This paper presents a colonoscope tracking method utilizing a colon shape estimation method.
CT colonography is used as a less-invasive colon diagnosis method.
If colonic polyps or early-stage cancers are found, they are removed in a colonoscopic examination.
In the colonoscopic examination, understanding where the colonoscope running in the colon is difficult.
A colonoscope navigation system is necessary to reduce overlooking of polyps.
We propose a colonoscope tracking method for navigation systems.
Previous colonoscope tracking methods caused large tracking errors because they do not consider deformations of the colon during colonoscope insertions.
We utilize the shape estimation network (SEN), which estimates deformed colon shape during colonoscope insertions.
The SEN is a neural network containing long short-term memory (LSTM) layer.
To perform colon shape estimation suitable to the real clinical situation, we trained the SEN using data obtained during colonoscope operations of physicians.
The proposed tracking method performs mapping of the colonoscope tip position to a position in the colon using estimation results of the SEN.
We evaluated the proposed method in a phantom study.
We confirmed that tracking errors of the proposed method was enough small to perform navigation in the ascending, transverse, and descending colons.
\end{abstract}



\section{Introduction}

We propose a colonoscope tracking method using a shape estimation network.
CT image-based diagnosis of the colon is performed as one of less invasive diagnosis methods.
It is called as CT colonography.
If colonic polyps or early-stage cancers are found in diagnosis, a colonoscopic examination or polypectomy is performed.
In a colonoscopic examination, a physician explores in the colon to find polyps.
However, understanding where the colonoscope running in the colon is difficult.
A colonoscope navigation system that navigates a physician to polyp positions during a colonoscope insertion is necessary.
To achieve a colonoscope navigation system, a colonoscope tracking method that finds the colonoscope position in the colon needs to be developed.

Colonoscope tracking methods were proposed by some research groups.
Image-based method\cite{Liu13} is difficult to keep tracking when unclear colonoscopic views appear.
Electromagnetic (EM) sensor will be necessary to perform tracking.
A system that shows colonoscope shape in the colon using EM sensors have been developed\cite{Ching10,Fukuzawa15}.
The system cannot perform navigation because it cannot map the colonoscope shape to the colon shape of a patient.
A colonoscope tracking method that utilizes a CT volume and EM sensors was proposed\cite{Oda17}.
This method obtains two curved lines representing colonoscope and colon shapes to estimate a colonoscope position on a CT volume coordinate system.
However, large tracking errors were caused due to colon deformations occurred during the colonoscope insertions.
To reduce this problem, estimation methods of the colon shape were proposed\cite{Odaspie18,Odamiccai18}.
The shape estimation network (SEN)\cite{Odamiccai18} accurately estimates colon shape utilizing temporal information.
The SEN performs estimation using a neural network containing long short-term memory (LSTM) layer.
Because the SEN employs a machine learning framework, it needs to be trained using data (colon and colonoscope shapes) obtained during colonoscope operations.
However, the SEN was trained using data measured during operations of a colonoscope by an engineering researcher.
The colonoscope operations were different from that of performed by physicians.

In this paper, we propose a colonoscope tracking method.
We utilize EM sensors attached to the colonoscope and a CT volume of a patient.
Colonoscope and colon shapes are obtained from them.
The SEN estimates a colon shape that are deformed by a colonoscope inserted to it.
To perform colon shape estimation based on real physician's data, we trained the SEN using data obtained during physician's colonoscope operations.
We perform colonoscope position estimation utilizing the colonoscope shape and the estimated colon shape obtained from the trained SEN.

\section{Colonoscope Tracking Method} \label{sec:method}

\subsection{Overview}

We obtain two shapes representing a colonoscope and a colon.
The colonoscope and colon shapes are registered to find a position in the CT volume coordinate system that corresponds to the colonoscope tip position.
The colon shape changes during colonoscope insertions.
We use the SEN\cite{Odamiccai18} to estimate deformed colon shapes.
Correspondences of points on the colonoscope shape and the estimated colon shape are established based on the deformation estimation results.
A colonoscope tip position in the CT volume coordinate system is calculated from the correspondence.

\subsection{Colonoscope and colon shapes}

\begin{figure}[tb]
\begin{center}
\includegraphics[width=0.8\textwidth, clip, trim=0 140 300 0]{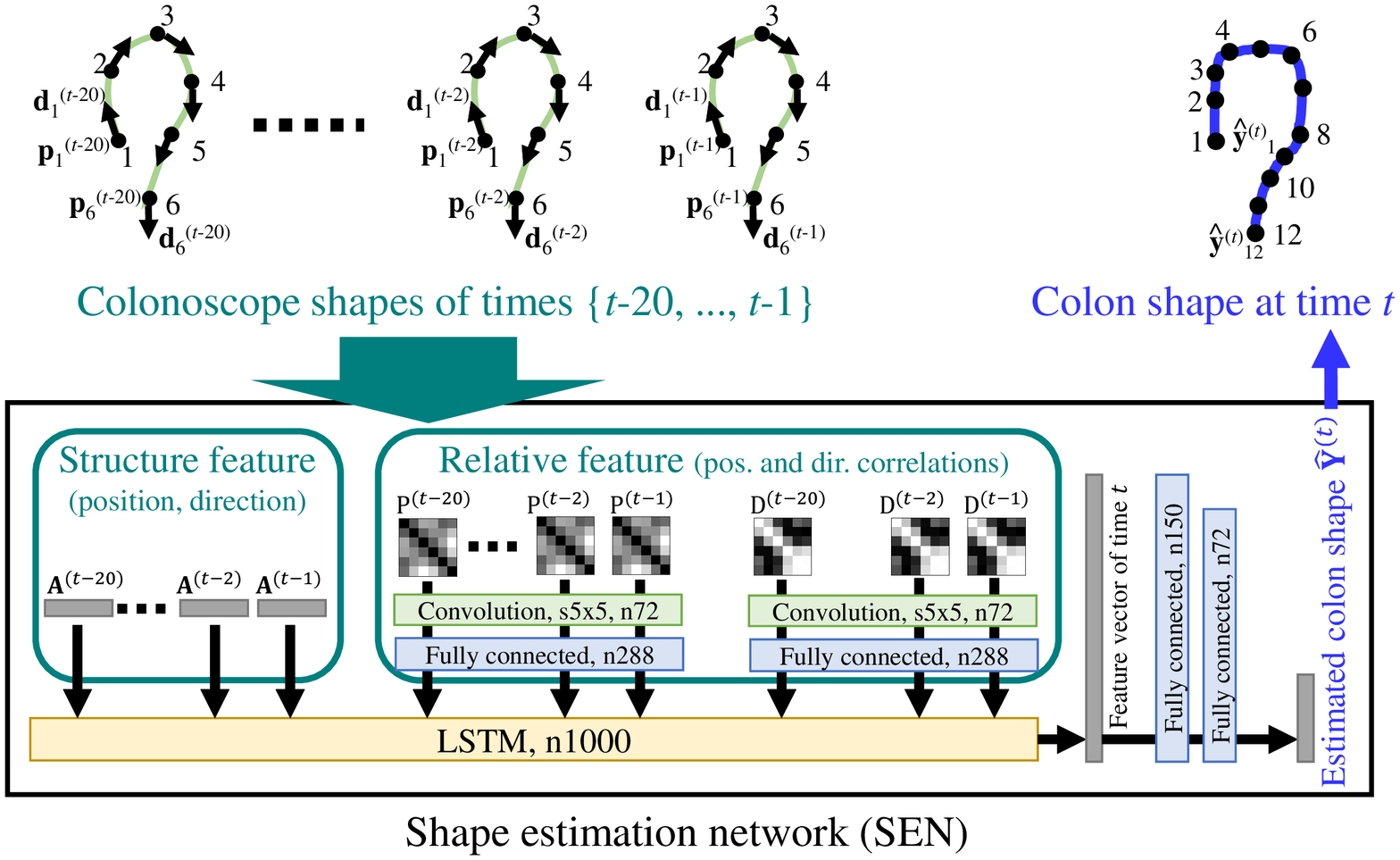}
\caption{SEN\cite{Odamiccai18} estimates colon shape at time $t$ from colonoscope shapes in time $\{t-20, \dots, t-1 \}$. ${\bf A}^{(t)}$ contains position and direction data of colonoscope shape. ${\rm P}^{(t)}$ and ${\rm D}^{(t)}$ are positional and directional correlations of colonoscope shape. In neural network layers, ``s5x5'' represents kernel size and ``n72'' represents number of kernel or unit.}
\label{fig:estimation}
\end{center}
\end{figure}

We represent the colonoscope and colon shapes as sets of points.
The colonoscope shape is a set of points and directions ${\bf X} = \left\{ {\bf p}_{n}, {\bf d}_{n}; n=1, \ldots, N \right\}$ aligned along a colonoscope.
$N$ is the total number of the points in the colonoscope shape.
The colon shape is a set of points ${\bf Y} = \left\{ {\bf y}_{m}; m=1, \ldots, M \right\}$ aligned along a colon.
$M$ is the total number of the points in the colon shape.
Examples of them are found in Fig. \ref{fig:estimation}.

\subsection{Colon shape estimation using SEN}

We use the SEN\cite{Odamiccai18} to estimate colon shapes which are deformed during colonoscope insertions.
The SEN is a neural network containing convolution and LSTM layers.
Its input is a set of the time-series colonoscope shapes.
It outputs an estimated colon shape ${\bf \hat{Y}} = \left\{ {\bf \hat{y}}_{m}; m=1, \ldots, M \right\}$ (Fig. \ref{fig:estimation}).

In the previous work\cite{Odamiccai18}, the SEN was trained using colonoscope and colon shapes measured during colonoscope operations performed by an engineering researcher.
We measured colonoscope and colon shapes during colonoscope operations by physicians.
The physician's data are used to train the SEN.
Measurement of the shapes are explained in the Sec. \ref{ssec:measurement}.

\subsection{Colonoscope tip position calculation}

A point ${\bf x}_{1}$ in the colonoscope shape represents the colonoscope tip position.
We find a point ${\bf \hat{y}}_{a}$ in the estimated colon shape which is closest to ${\bf x}_{1}$.
We find a point ${\bf y}_{a}$ in the colon shape that corresponds to ${\bf \hat{y}}_{a}$.
Each point in the colon and estimated colon shapes have one-to-one correspondence.
As a result of these two-steps mappings, we obtain ${\bf y}_{a}$ as a point in the CT volume coordinate system where the colonoscope tip located.

\section{Experiments} \label{sec:experiments}

\subsection{Materials}

\begin{figure}[tb]
\begin{center}
\begin{tabular}{cc}
\includegraphics[width=0.65\textwidth, clip, trim=0 120 300 0]{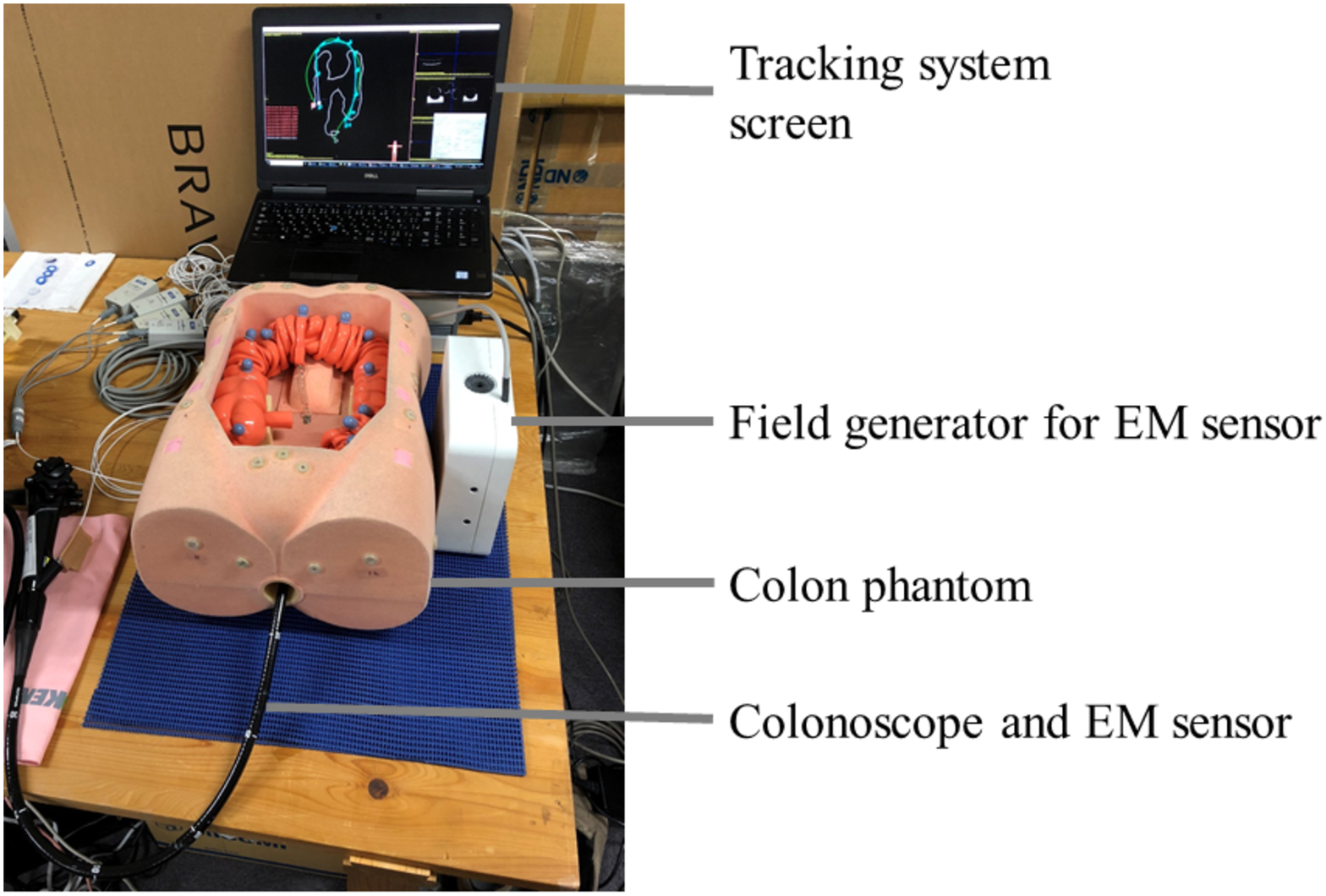} & 
\includegraphics[width=0.3\textwidth, clip, trim=0 120 650 0]{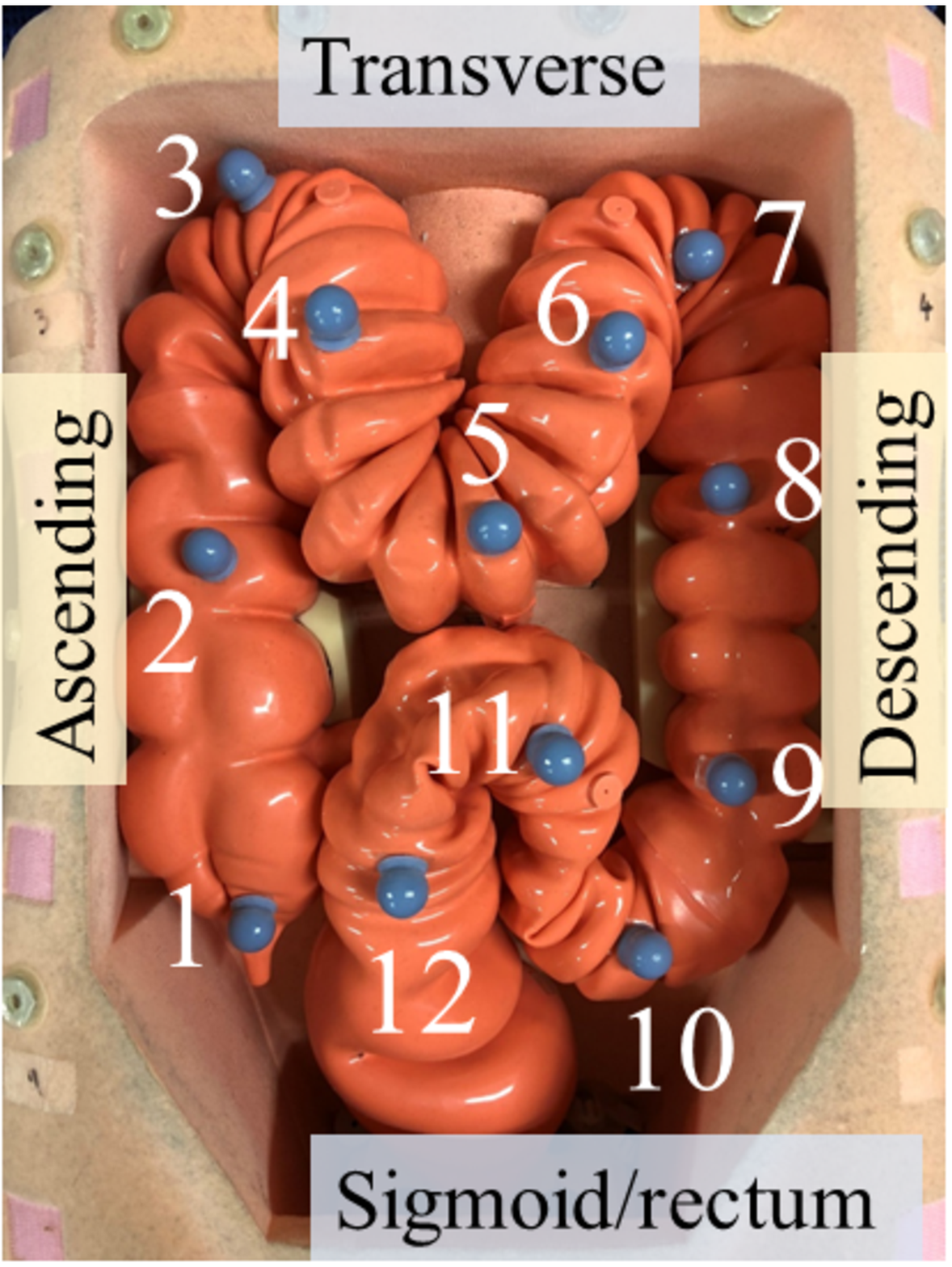}
\\
(a) & (b)
\end{tabular}
\end{center}
\caption{(a) experimental materials. (b) positions of 12 markers on the colon phantom.}
\label{fig:devices}
\end{figure}

We evaluated the proposed colonoscope tracking method in phantom-based experiments.
We use a colon phantom (colonoscopy training model type I-B, Koken, Tokyo, Japan), a CT volume of the phantom, a colonoscope (CF-Q260AI, Olympus, Tokyo, Japan), an EM sensor (Aurora 5/6 DOF Shape Tool Type 1, NDI, Ontario, Canada), and a distance image sensor (Kinect v2, Microsoft, WA, USA).
Experimental materials are shown in Fig. \ref{fig:devices} (a).

\subsection{Colonoscope and colon shapes measurement}
\label{ssec:measurement}

The colonoscope shape is measured using the EM sensor.
The EM sensor is strap-shaped with six sensors at its tip and points along its body.
It is inserted into the colonoscope working channel.
The colonoscope shape is a set of six points.

The colon shape is measured using the distance image sensor.
We obtain 12 three-dimensional points on the colon phantom using the sensor.
The points are aligned along the colon centerline.
The colon shape is a set of the 12 points.

Both of the colonoscope and colon shapes are registered to the CT volume coordinate system.
The colonoscope shape is registered using the iterative closest point (ICP) algorithm\cite{Besl92} 
by minimizing distance between the colonoscope shape and a point set on the colon centerline in the CT volume.
The colon shape is registered manually.

\begin{figure}[tb]
\begin{center}
\includegraphics[width=0.9\textwidth, clip, trim=0 50 50 0]{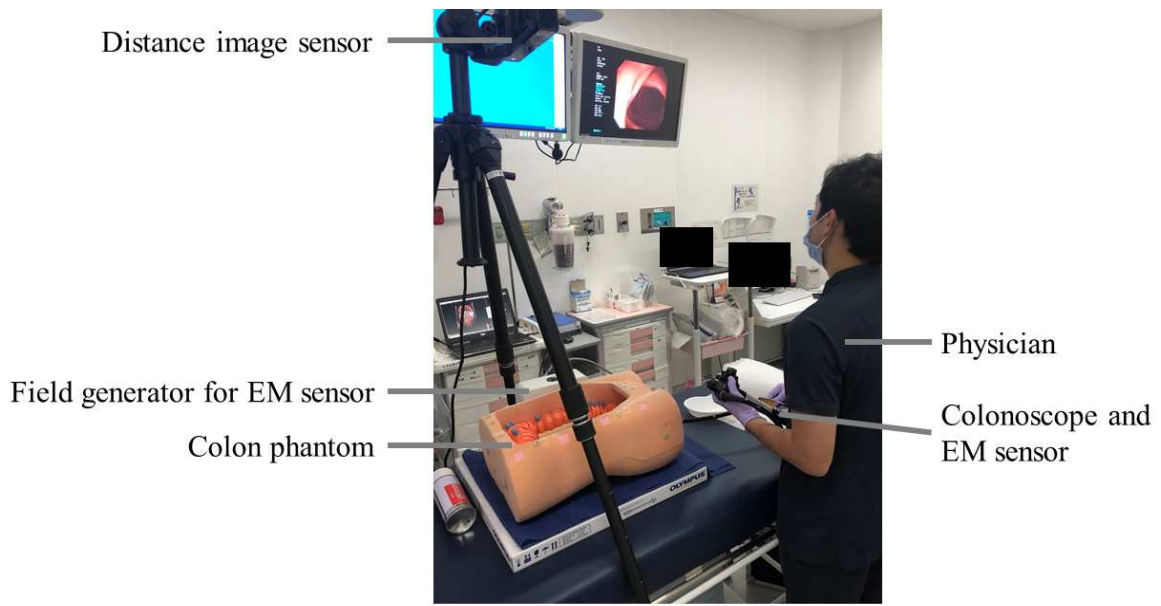}
\caption{Experimental setup for measuring colonoscope operations of physicians. EM sensor and distance image sensor measure colonoscope and colon shapes during colonoscope insertions by physicians.}
\label{fig:experiment_pysician}
\end{center}
\end{figure}

To obtain training data of the SEN\cite{Odamiccai18}, we measured the colonoscope and colon shapes during colonoscope insertions to the colon phantom.
Operators moved the colonoscope tip from the cecum to anus.
10 times of colonoscope movements from the cecum to anus were measured in our experiments.
Two physicians (four times colonoscope movements) in the Nagoya University Hospital and an engineering researcher (six time colonoscope movements) were operated.
The SEN was trained using the measured data.
Figure \ref{fig:experiment_pysician} shows the configuration of experimental setup for measurement of physician's colonoscope operation.

\subsection{Definition of tracking error}

We placed 12 markers on the surface of the colon phantom (Fig. \ref{fig:devices} (b)).
The markers are identifiable points both on the colon phantom and in its CT volume.
We employed tracking error as a criterion of tracking performance evaluation.
The tracking error is the distance between an estimated colonoscope tip position and a marker position in the CT volume, when the colonoscope tip comes closest to the marker.

\subsection{Tracking and evaluation}

In colonoscopic examinations, physicians observe the colon while retracting the colonoscope after its insertion up to the cecum.
We assume the colonoscope tip is inserted up to the cecum when the colonoscope tracking starts.
The colonoscope tracking is performed while retracting the colonoscope from the cecum to the anus.

We performed colonoscope tracking five times.
An engineering researcher operated the colonoscope.

\section{Results}

\begin{figure}[tb]
\begin{center}
\includegraphics[width=0.6\textwidth, clip, trim=0 150 400 0]{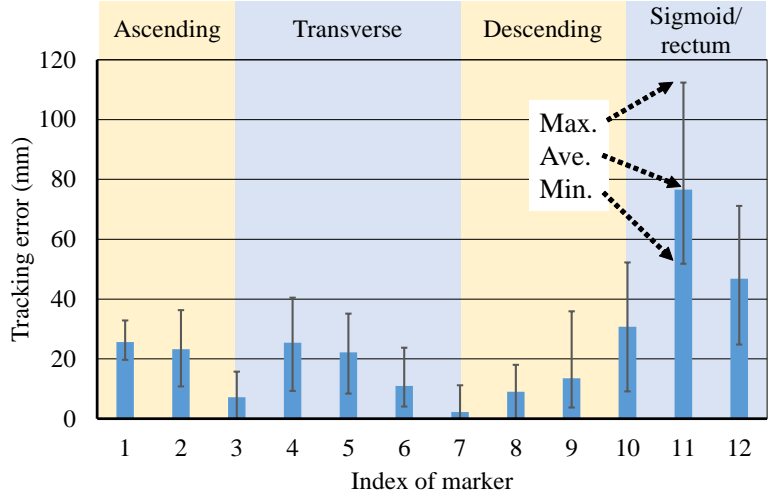}
\caption{Tracking error at each marker. Bar graph shows average tracking error with maximum and minimum values.}
\label{fig:error}
\end{center}
\end{figure}

Average, maximum, and minimum tracking errors of five times colonoscope tracking are shown in Fig. \ref{fig:error}.
Tracking errors were small in the ascending, transverse, and descending colons.
Large tracking errors were observed in the sigmoid colon and rectum.

\section{Discussion}

A physician who specializes in gastroenterology commented that tracking errors smaller than 50 mm are acceptable for colonoscope navigation to polyp positions.
This is because polyps are observable from the colonoscope camera when positions of the colonoscope tip and polyps are closer than 50 mm.
The proposed method achieved average tracking errors smaller than 50 mm in the ascending, transverse, and descending colons.
The previous method\cite{Oda17} achieved it in the ascending and descending colons (under setting of $R=40$ mm).
In the transverse colon, smaller tracking errors were observed in the proposed method.
The previous method\cite{Oda17} does not consider colon deformation.
It makes tracking errors in the transverse colon large because the transverse colon is easy to deform.
Consideration of colon deformation strongly contributed to reduce tracking errors.

The tracking errors in the sigmoid colon and rectum were large.
This is because the sigmoid colon and rectum have heavy deformation during the colonoscope insertions.
The deformation contains considerable variation.
The data used to train the SEN was not enough to provide information about deformation in these colon regions.
We need to correct data that containing various deformations of the sigmoid and rectum to improve both of the colon shape estimation and tracking accuracies.



\section{Conclusions}

We proposed a colonoscope tracking method using the SEN.
Previous tracking method caused large tracking errors due to colon deformations.
In our method, the SEN estimates deformed colon shape.
The SEN was trained using data obtained during colonoscope operations by physicians.
The estimated colon shape is used to make correspondence between the colonoscope tip position and its position in the colon.
In our phantom study, tracking errors in the transverse colon were reduced by considering colon deformations.
Future work includes evaluation using more markers, reduction of tracking error in the sigmoid and rectum, and application to tracking in human colon.

\acknowledgments 
 
Parts of this research were supported by the MEXT/JSPS KAKENHI Grant Numbers 26108006, 17H00867, the JSPS Bilateral International Collaboration Grants, and the JST ACT-I (JPMJPR16U9).





\bibliography{19spie_abstract_cite}   
\bibliographystyle{spiebib}   

\end{document}